\documentclass[letterpaper, 10 pt, conference]{ieeeconf}  %
\IEEEoverridecommandlockouts                              

\usepackage{graphics} 
\usepackage{epsfig} 
\usepackage{mathptmx} 
\usepackage{times} 
\usepackage{amsmath} 
\usepackage{amssymb}  
\usepackage{algorithm}
\usepackage{algorithmic}
\usepackage{multicol}
\usepackage{multirow}
\usepackage{wasysym}
\usepackage{booktabs}
\usepackage{caption}
\usepackage{subcaption}
\usepackage{sidecap}
\usepackage{pifont}
\usepackage{xcolor}
\usepackage[colorlinks=true,linkcolor=red,urlcolor=blue,citecolor=green,anchorcolor=blue]{hyperref}
\definecolor{ube}{rgb}{0.82, 0.62, 0.91}
\newcommand{\cmark}{\ding{51}}%
\newcommand{\xmark}{\ding{55}}%
\newcommand\mypar[1]{\par\vspace{1.0mm}\noindent\textbf{#1}\;\;}
\title{\LARGE \bf
Streaming Motion Forecasting for Autonomous Driving
}

\author{Ziqi Pang$^1$, Deva Ramanan$^2$, Mengtian Li$^2$, Yu-Xiong Wang$^1$ \thanks{$^1$University of Illinois Urbana-Champaign, $^2$Cargenie Mellon University \newline Corresponding to {\tt{\{ziqip2, yxw\}@illinois.edu }}}}

\begin{document}
\maketitle
\thispagestyle{empty}
\pagestyle{empty}
\begin{abstract}
Trajectory forecasting is a widely-studied problem for autonomous navigation. However, existing benchmarks evaluate forecasting based on independent \emph{snapshots} of trajectories, which are not representative of real-world applications that operate on a \emph{continuous stream} of data. To bridge this gap, we introduce a benchmark that continuously queries future trajectories on streaming data and we refer to it as ``streaming forecasting.'' Our benchmark inherently captures the disappearance and re-appearance of agents, presenting the emergent challenge of \emph{forecasting for occluded agents}, which is a safety-critical problem yet overlooked by snapshot-based benchmarks. Moreover, forecasting in the context of continuous timestamps naturally asks for \emph{temporal coherence} between predictions from adjacent timestamps. Based on this benchmark, we further provide solutions and analysis for streaming forecasting. We propose a plug-and-play meta-algorithm called ``\emph{Predictive Streamer}'' that can adapt any snapshot-based forecaster into a streaming forecaster. Our algorithm estimates the states of occluded agents by propagating their positions with multi-modal trajectories, and leverages differentiable filters to ensure temporal consistency. Both occlusion reasoning and temporal coherence strategies significantly improve forecasting quality, resulting in 25\% smaller endpoint errors for occluded agents and 10-20\% smaller fluctuations of trajectories. Our work is intended to generate interest within the community by highlighting the importance of addressing motion forecasting in its \emph{intrinsic} streaming setting. Code is available at \href{https://github.com/ziqipang/StreamingForecasting}{https://github.com/ziqipang/StreamingForecasting}.

\end{abstract}

\section{INTRODUCTION}

\begin{figure}
    \centering
    \begin{subfigure}[tb]{1.0\linewidth}
    \vspace{-2mm}
    \caption{Streaming Forecasting}
    \label{fig:teaser_streaming}
    \includegraphics[width=1.0\linewidth]{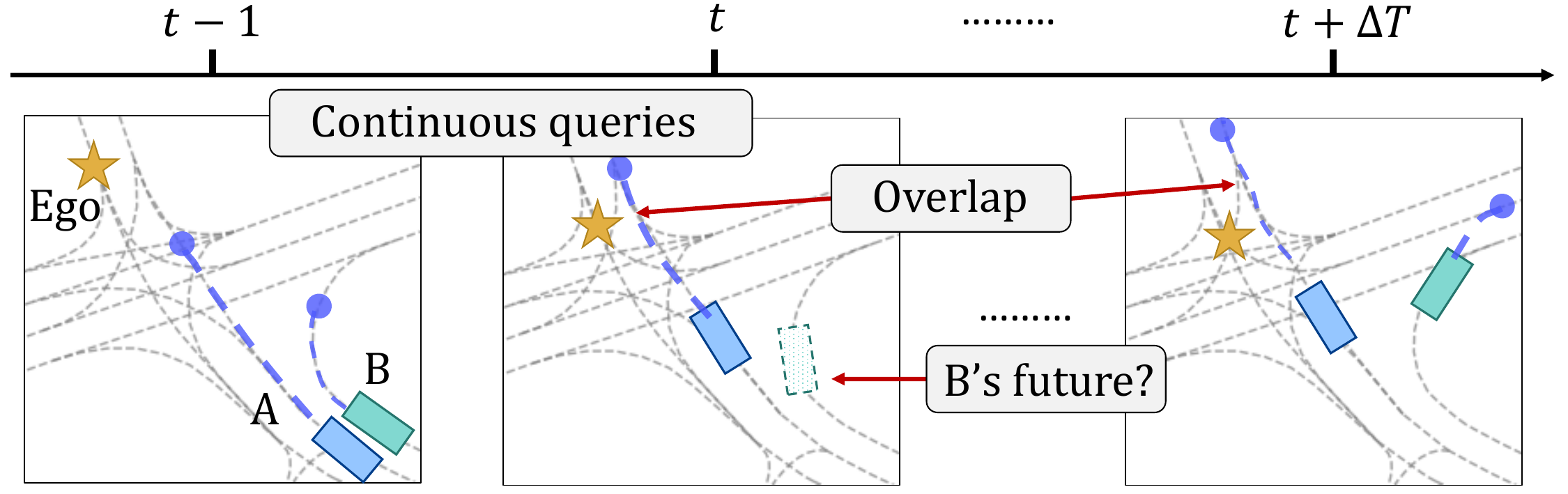}
    \end{subfigure}
    \begin{subfigure}[tb]{1.0\linewidth}
    \caption{Snapshot-Based Forecasting}
    \label{fig:teaser_snapshot}
    \includegraphics[width=1.0\linewidth]{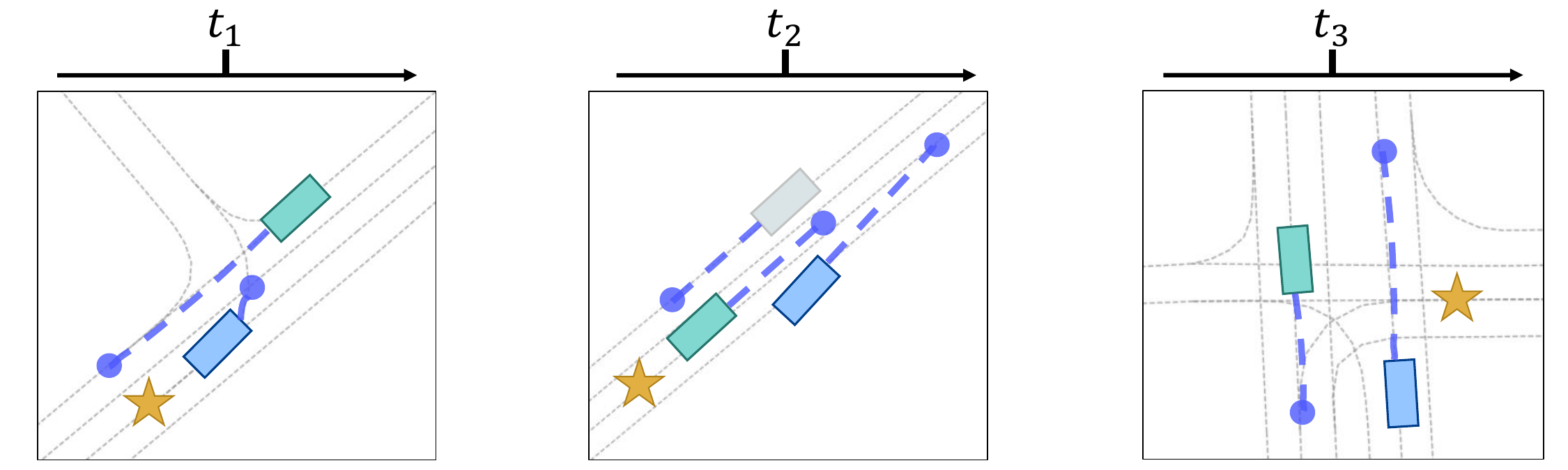}
    \end{subfigure}
    \caption{(\textbf{Best viewed in color}) Gray lines: HD-Map; Blue lines: predicted trajectories. \textbf{(a)} Our streaming setup queries predictions on \emph{continuous frames} to reflect the streaming property in the real world. Under our setting, the challenge of \emph{forecasting for occluded agents} emerges from the frequent disappearance and re-appearance of agents (\emph{e.g.}, agent B, highlighted by ``B's future''), and \emph{temporal coherence} becomes a natural constraint for predictions in adjacent timestamps (highlighted by ``Overlap''). \textbf{(b)} The snapshot-based setup in existing forecasting benchmarks sample queries \emph{independently} that are isolated in space and time. Such a setting hides away \emph{occlusion} and \emph{temporal coherence} challenges, which are presented in realistic streaming forecasting.}
    \vspace{-2mm}
    \label{fig:teaser}
\end{figure}

Motion forecasting is \emph{inherently} a \emph{streaming} task for autonomous driving, as it operates on continuous data streams. Imagine agents constantly moving in a dynamic traffic scene, as illustrated in Fig.~\ref{fig:teaser_streaming}. Naturally, agents may temporarily disappear due to \emph{occlusions} and then re-appear, such as agent B. Even when an agent is invisible, a forecasting algorithm or forecaster is still supposed to predict its future trajectories. Ignoring occlusions could lead to the sudden appearance of agents, posing critical safety risks. Moreover, in this streaming setting where trajectories predicted on continuous timestamps have overlaps, ensuring \emph{temporal coherence} between the trajectories becomes a key challenge. The forecaster needs to produce stable and smooth trajectories over time to serve downstream planning and control models.

Unfortunately, current motion forecasting benchmarks universally adopt a \emph{snapshot}-based setup~\cite{caesar2020nuscenes}\cite{chang2019argoverse}\cite{ettinger2021large}, as illustrated in Fig.~\ref{fig:teaser_snapshot} -- it only considers \emph{independent} and \emph{isolated} snapshots of trajectories, without explicitly modeling the spatial-temporal connection among them. Such a setting standardizes but over-simplifies the motion forecasting problem, making it not representative of the streaming nature of the real world. 

To overcome this limitation and demonstrate the streaming reality of autonomous driving, we propose ``\emph{streaming forecasting}'' which involves \emph{continuously} querying the future trajectories of agents \emph{on every frame}. Such a new perspective of motion forecasting presents a clear departure from the snapshot-based setting with independent samples. Importantly, our formulation also exhibits two novel real-world challenges. (1) \emph{Forecasting for occluded agents}: as in frame $t$ of Fig.~\ref{fig:teaser_streaming}, our streaming setup captures the changing visibility of agents and reveals the occlusion challenge. However, snapshot-based datasets~\cite{caesar2020nuscenes}\cite{chang2019argoverse}\cite{ettinger2021large} uniformly assume that the target agents are visible and collect the data accordingly. (2) \emph{Temporal coherence}: as in frames $t$ and $t+\Delta T$ of Fig.~\ref{fig:teaser_streaming}, trajectories in neighboring frames have overlaps and our streaming setting reflects such a constraint by modeling motion forecasting continuously, which is impossible for isolated samples in the snapshot-based setting.

Based on this formulation, we introduce a new benchmark for streaming forecasting. Our key insight is to \emph{re-purpose tracking} datasets, which offer realistic observations from the ego vehicle. This strategy mitigates the bias of forecasting datasets in their data collection and processing. Specifically, Argoverse~\cite{chang2019argoverse} is chosen for supporting modern vectorized high-definition maps (HD-Maps) on its tracking dataset. We call our benchmark \emph{Argoverse-SF}, which faithfully captures the challenges of frequent occlusions and temporal continuity of predictions. To evaluate the performance of forecasting algorithms in this streaming setting, we design tailored metrics that investigate occlusion reasoning and temporal coherence.

To address both challenges, we further propose a simple yet effective meta-algorithm, called ``\emph{Predictive Streamer},'' to adapt any existing snapshot-based forecaster into the streaming world. Our predictive streamer estimates the states of occluded agents to enable robust predictions for them. While inferring occluded positions is a long-standing problem in 3D perception~\cite{simpletrack}\cite{pang2023standing}\cite{wang2021immortal}, existing strategies of using the Kalman filter or single-modal trajectories are insufficient for the forecasting purpose, because the forecaster depends on the estimated occluded positions and amplifies the errors in occlusion reasoning. We discover that a more advanced strategy of utilizing \emph{multi-modal trajectories} predicted by the forecaster performs better, as it provides higher-quality estimation covering a larger spectrum of motion distributions.

For \emph{temporal coherence}, our predictive streamer introduces \emph{differentiable filters} (DFs)~\cite{haarnoja2016backprop} into multi-modal trajectory forecasting, adapting them from the state estimation literature. Our modified DFs represent future trajectories as hidden states and emphasize the temporal coherence of multi-modal trajectories in the process model of DFs. Compared with recurrent models, \emph{e.g.}, long short-term memory networks (LSTMs), our DFs also perform better because of the explicit modeling of temporal continuity. Finally, our approach enhances the temporal coherence of trajectories and improves the accuracy of forecasting results accordingly.

In brief, we have made the following contributions.
\begin{enumerate}[leftmargin=*, noitemsep, nolistsep]
    \item \textbf{\emph{Benchmark.}} \textbf{(a)} We introduce a novel \emph{streaming forecasting} formulation and construct an associated ``Argoverse-SF'' benchmark by re-purposing the tracking dataset. \textbf{(b)} Our benchmark reveals and captures the inherent challenges in a streaming world: ``\emph{occlusion reasoning}'' and       ``\emph{temporal coherence},'' which are previously ignored by the widely-used snapshot-based benchmarks.
    \item \textbf{\emph{Solution.}} \textbf{(a)} We propose a plug-and-play meta-algorithm called ``\emph{Predictive Streamer}'' that can adapt any snapshot-based forecaster into a streaming forecaster. \textbf{(b)} We instantiate our meta-algorithm by applying \emph{multi-modal trajectories} to estimate occluded positions and introducing \emph{differentiable filters} to improve the temporal coherence of multi-modal trajectories. Our solution addresses the two key challenges in streaming forecasting, decreasing the minimum final displacement error (minFDE) by 25\% for occluded agents and reducing the fluctuations of trajectories by 10-20\%. 
\end{enumerate}
\section{RELATED WORK}

\subsection{Motion Forecasting in Autonomous Driving}
Motion forecasting methods aim to predict the trajectories of agents by modeling their dynamics, interactions, and relationship with high-definition maps (HD-Maps). Previous work has developed effective encoders to extract agent features using graph neural networks (GNNs)~\cite{gao2020vectornet}\cite{liang2020learning}, LSTMs~\cite{salzmann2020trajectron++}, and transformers~\cite{liu2021multimodal}\cite{ngiam2021scene}\cite{yuan2021agentformer}. Other work has focused on decoding trajectories from agent features using improved sampling strategies and learning objectives~\cite{gilles2021home}\cite{gilles2022gohome}\cite{gu2021densetnt}\cite{varadarajan2022multipath++}\cite{zhao2021tnt}. Despite rapid progress, they are all developed and evaluated under the contrived snapshot-based forecasting benchmarks~\cite{caesar2020nuscenes}\cite{chang2019argoverse}\cite{ettinger2021large}. Although some work~\cite{salzmann2020trajectron++}\cite{zhou2023query} provides interfaces for making predictions on continuous timestamps, we found that adding recurrent neural networks or relative positional embedding cannot fully address the streaming forecasting challenges, especially occlusion reasoning. Instead, we analyze the limitations of existing models and propose simple yet effective strategies to estimate occluded positions and improve temporal coherence. In addition, our streaming forecasting is broadly relevant to \emph{open-loop planning}~\cite{caesar2021nuplan} which similarly models a streaming world, with the key difference of predicting the motion of the ego-vehicle, while we focus on surrounding agents. 

\subsection{Joint Perception and Forecasting}

Motion forecasting is often studied jointly with other perception modules. This naturally positions forecasting in data streams. End-to-end perception and prediction connects forecasting with 3D perception~\cite{casas2021mp3}\cite{gu2022vip3d}\cite{fiery}\cite{luo2018fast}\cite{pang2023standing}, which mainly explores how to benefit motion forecasting from the 3D perception features instead of improving the capability of forecasting models. By contrast, our work analyzes the setbacks of existing forecasting models under a streaming formulation and proposes an advanced ``predictive streamer'' to address the challenges of occlusion reasoning and temporal coherence.

\subsection{Differentiable Filters}

Differentiable filters (DFs)~\cite{abbeel2005discriminative}\cite{haarnoja2016backprop} combine the benefits of neural networks with Bayes filters~\cite{kalman1960new}, where a neural network handles high-dimensional sensor inputs and enables the filters with the flexibility to adapt to a wide range of scenarios. Recently, DFs have demonstrated their effectiveness in 3D tracking and visual odometry~\cite{jonschkowski2018differentiable}\cite{karkus2018particle}\cite{kloss2021train}\cite{yi2021differentiable}, as well as robot manipulation~\cite{lee2020multimodal}\cite{zachares2021interpreting}. However, they mainly concentrate on state estimation and apply filters to \emph{current} positions of objects. Instead, we extend DFs to enhance the prediction of \emph{long-term future} trajectories of agents in autonomous driving scenarios, which exhibit more significant stochastic fluctuations and multi-modalities.
\section{STREAMING FORECASTING BENCHMARK}

\subsection{Background: Conventional Motion Forecasting}
\label{sec:background}

\begin{figure}
    \centering
    \includegraphics[width=1.0\linewidth]{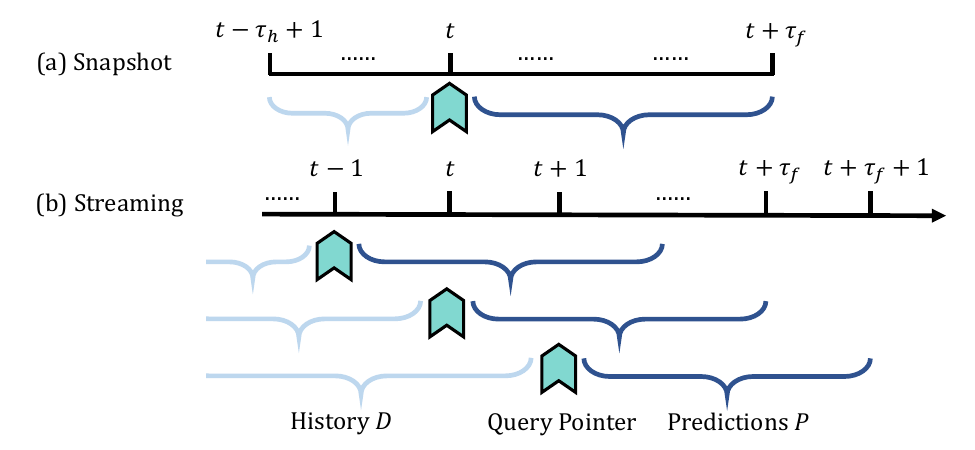}
    \vspace{-2mm}
    \caption{Formulations of \emph{snapshot}-based and \emph{streaming} forecasting. \textbf{(a)} Conventional forecasting handles snapshots with a fixed length and is only intended for single-timestamp predictions. \textbf{(b)} Our \emph{streaming} setup queries the future trajectories on \emph{every} timestamp, aligning better with the streaming world. Compared with the snapshot-based setting, the length of history grows without an upper bound.}
    \vspace{-4mm}
    \label{fig:formulation}
\end{figure}

\mypar{Motion forecasting.} Given historical observations, motion forecasting aims to predict the trajectories of agents in future timestamps. In the conventional formulation (Fig.~\ref{fig:formulation}\textcolor{red}{a}), historical observations have a fixed length of $\tau_h$, and consist of the center positions of $N$ agents $C_{t-\tau_h+1:t}=\{ c^i_{t-\tau_h+1:t} \}_{i=1}^N$ and an HD-Map $M$. The future trajectories are multi-modal predictions denoted as $P_{t:t+\tau_f}=\{\{p^{i,k}_{t:t+\tau_f}\}_{k=1}^K\}_{i=1}^N$, where $K$ is the number of predictions and each $p^{i, k}_{t_1:t_2}$ represents the estimated \emph{movement} between timestamps $t_1$ and $t_2$.

\mypar{Snapshot-based benchmarks.} As illustrated in Fig.~\ref{fig:teaser_snapshot}, existing benchmarks, such as~\cite{caesar2020nuscenes}\cite{chang2019argoverse}\cite{ettinger2021large}, are typically constructed by collecting \emph{independent} and \emph{isolated} ``snapshots'' of trajectories, without explicitly considering the spatial-temporal connection among them. Within each snapshot, only the predictions for a \emph{single} timestamp are queried. 

\mypar{Forecasters.} Most of the current forecasting models are based on encoder-decoder architectures. The encoder takes as input the past locations $C_{t-\tau_h+1:t}$ and the HD-Map $M$ to generate agent features $F_t$. The decoder then predicts the trajectories $P_{t:t+\tau_f}$ from the agent features $F_t$. That is,
\begin{equation}
    \label{eqn:enc_dec}
    F_{t}=\mathtt{Encoder}(C_{t-\tau_h+1:t}, M),\quad P_{t:t+\tau_f}=\mathtt{Decoder}(F_{t}).
\end{equation}
We denote a snapshot-based forecasting model as $\mathtt{Model}=\{\mathtt{Encoder}, \mathtt{Decoder}\}$.

\subsection{Our Formulation}
\label{sec:streaming_formulation}
Although it is widely used, the snapshot-based formulation is incompatible with the ``streaming'' nature of the real world. We thus propose a \emph{streaming forecasting} formulation that evaluates a forecasting algorithm on continuous timestamps instead of a single run. As shown in Fig.~\ref{fig:formulation}\textcolor{red}{b}, streaming forecasting iteratively executes two steps: (1) \emph{Input} the positions of agents; (2) \emph{Query} the corresponding future trajectories.

\mypar{Input.} We define $A_t$ as the set of IDs of all the agents that have ever appeared up to time $t$ (details of agent selection are in Sec.~\ref{sec:benchmark}). At time $t$, the input consists of the 3D coordinates $c_t^a$ and visibility $v^a_t$ for every agent $a\in A_t$. If agent $a$ is visible or present at time $t$, we set $v^a_t$ as true; otherwise, we set $v^a_t$ as false and $c_t^a \xleftarrow{} \diameter$. We represent all the input data at time $t$ as $D_t=\{(c_t^a, v_t^a)\}_{a\in A_t}$, and $D_{t-\tau_h+1:t}$ denotes all the input data between frames $t-\tau_h+1$ and $t$.

\mypar{Query.} Streaming forecasting queries the future trajectories of agents in $A_t$ on every timestamp. We adopt the same multi-modality setting with $K$ futures for each agent as the conventional formulation in Sec.~\ref{sec:background}. Thus, the predictions for agent $a\in A_t$ are $P_{t:t+\tau_f}^a=\{{p^{a, k}_{t:t+\tau_f}\}_{k=1}^K}$. The corresponding ground truth is $G_{t:t+\tau_f}=\{g^{a}_{t:t+\tau_f}\}_{a\in A_t}$, which comes from the observations in future frames $D_{t:t+\tau_f}$ of the streaming data.%

\mypar{Evaluated agents.} Our streaming forecasting queries and evaluates all the agents in $A_t$, further setting it apart from conventional forecasting benchmarks that only concern a subset of target agents. While selecting specific target agents may help to focus on interesting movements, pre-assuming target agents can be risky in safety-critical applications. For similar considerations, joint perception and forecasting studies~\cite{casas2021mp3}\cite{gu2022vip3d}\cite{luo2018fast}\cite{pang2023standing} often evaluate all the agents.

By evaluating all the agents, our streaming forecasting can capture the challenge of ``\emph{forecasting for occluded agents}.'' By contrast, previous forecasting benchmarks implicitly assume the visibility of target agents and overlook this issue. Note that occluded agents also have available ground truth for future trajectories in streaming forecasting, because \emph{they can re-appear after occlusion}. This makes streaming forecasting a natural scenario for investigating occlusions and addressing the challenges they pose.

\subsection{Argoverse-SF Benchmark}
\label{sec:benchmark}
Guided by our formulation, we construct the Argoverse-SF benchmark for streaming forecasting. We make the following important design choices and contributions.

\mypar{Using tracking instead of forecasting data.} Autonomous driving datasets \cite{caesar2020nuscenes}\cite{chang2019argoverse}\cite{ettinger2021large} typically have separate splits for tracking and forecasting. While using the forecasting data seems the obvious choice, such data have been processed for snapshot-based evaluation: predicting on a single timestamp instead of continuous timestamps, and filtering out noisy (\emph{e.g.}, occluded) agents for a cleaner setup. By contrast, tracking data satisfy our purpose, as the input to forecasting components comes from an upstream tracker in autonomous driving. Therefore, tracking data are a more appropriate option for our benchmark to reflect the real world faithfully. 

\mypar{Argoverse vs. other datasets.} We prioritize evaluating datasets that support algorithms based on modern vectorized HD-Maps. As Waymo~\cite{ettinger2021large} does not include HD-Maps in its tracking data, Argoverse~\cite{chang2019argoverse} and nuScenes~\cite{caesar2020nuscenes} become our main options. We ultimately choose Argoverse, because more forecasters using vectorized HD-Maps are studied on Argoverse than nuScenes.

\mypar{Agent positions.} As for the agent positions in our benchmark, instead of relying on the tracking results from a specific tracker, we leverage high-quality ground truth from Argoverse to create an ``oracle'' tracker. Specifically, we use the centers of ground truth bounding boxes as the input positions. This design avoids the influence of specific trackers and enables us to concentrate on the difficulties of forecasting. For instance, the occlusions in our Argoverse-SF are faithful in that they reflect the actual invisibility of occluded agents and cannot be filled in by human annotators. We further control the set of agents $A_t$ by using the oracle life-cycle management~\cite{simpletrack}\cite{wang2021immortal}: an agent is removed from the set of agents only if it does not re-appear. 

\mypar{Selection of agents.} For a realistic setup and fair assessment of forecasting algorithms, Argoverse-SF carefully selects the agents. First, agents outside the perception range ($>$100m) or far from the road (outside the regions of interest in Argoverse) are excluded. Second, Argoverse-SF begins querying at $\tau_h=20$ frame, ensuring that the benchmark does not create short histories artificially. Finally, Argoverse-SF only includes automobile agents such as vehicles, buses, and trailers, aligning with the Argoverse forecasting dataset, and we leave the implementation of additional agent types and datasets as future work.

\begin{table}[tb]
\caption{The number of queried predictions (``Queries'') and the total number of agents to which these predictions belong (``Agents'') in Argoverse-SF.}\vspace{-2mm}
	\begin{subtable}[h]{0.99\linewidth}
 \caption{Statistics of Argoverse-SF training set.}\vspace{-2mm}
        \centering
        \resizebox{1.0\linewidth}{!}
        {
        \begin{tabular}{c|cccc}
        \toprule
          & Moving-Visible & Moving-Occluded & Static-Visible & Static-Occluded \\
        \midrule
        Queries & 90814 & 8342 & 103838 & 20875 \\
        \midrule
        Agents & 1042 & 520 & 2424 & 1494 \\
        \bottomrule
       \end{tabular}
       }
    \end{subtable}
    \begin{subtable}[h]{0.99\linewidth}
    \caption{Statistics of Argoverse-SF validation set.}\vspace{-2mm}
        \centering
        \resizebox{1.0\linewidth}{!}
        {
        \begin{tabular}{c|cccc}
        \toprule
          & Moving-Visible & Moving-Occluded & Static-Visible & Static-Occluded \\
        \midrule
        Queries & 29124 & 2968 & 47957 & 5069 \\
        \midrule
        Agents & 350 & 187 & 742 & 451 \\
        \bottomrule
       \end{tabular}
       }
    \end{subtable}
    \label{tab:distribution}
    \vspace{-2mm}
\end{table}

\mypar{Dataset splits.} Argoverse-SF adopts the same training and validation sets as Argoverse's tracking dataset: 65 and 24 sequences for training and evaluation, respectively. We use all the frames in the logs of Argoverse, which are 10Hz sensor inputs with a duration of 15-30 seconds. The training and validation sets have 9,937 and 3,839 timestamps with valid queries of future trajectories, respectively. Note that Argoverse-SF allows using Argoverse's forecasting dataset for pre-training forecasters. This is useful, because the forecasting dataset contains a curated collection of diversified trajectories that are necessary for learning effective initialization and comparing fairly with existing models.

\mypar{Evaluation metrics.} Argoverse-SF quantitatively evaluates the performance of the predictions $P_{t:t+\tau_f}$ by comparing them with the ground truth $G_{t:t+\tau_f}$, which is the observations of agents' positions in future frames $D_{t:t+\tau_f}$ (Sec.~\ref{sec:streaming_formulation}). We adopt the commonly-used evaluation metrics of minimum final displacement error (minFDE), minimum average displacement error (minADE), and miss rate (MR). However, due to the frequent occlusion and de-occlusion of agents in streaming data, some frames and agents may not have ground truth. Therefore, we only compute the metric values for timestamps with available ground truth, \emph{i.e.}, where the visibility mask $v_t^a$ is true. For example, the minFDE for agent $a$ is calculated as: 
\begin{equation}
    \label{eqn:metrics}
    \text{minFDE}(a) = \frac{1}{|T^a|}\sum_{t\in T^a} \text{minFDE}(\{p^{a,k}_{t:t+\tau_f}\}_{k=1}^K, g^a_{t:t+\tau_f}),
\end{equation}
where $T^a$ is the set of all the queried timestamps with valid ground truth for agent $a$: $T^a=\{ t|v^a_{t+\tau_f}\ \text{is true} \}$. For metrics like minADE, we adopt a ``masked'' version by performing average only on the frames with $v^a_t$ being true.

\mypar{Breakdown into subsets for evaluation.} Conventional motion forecasting only considers moving and visible agents. By contrast, our streaming forecasting introduces three new types of agents: moving-occluded, static-visible, and static-occluded, and we evaluate all of them. Table~\ref{tab:distribution} summarizes the statistics with respect to the number of agents and queries, showing that the four types (subsets) of agents have a highly-imbalanced distribution. Here, we intuitively define \emph{moving} agents as those having an overall traveling distance greater than 3.0 meters. To avoid the domination of a single motion pattern or visibility, we introduce the four-subset division of movements and visibility for evaluation. Specifically, we first average the metrics within each subset $A$ of the agents, calculated as ``$\text{minFDE}(A) = (1/|A|) \sum_{a\in A}(\text{minFDE}(a))$,'' and then average the metrics across the four subsets to evaluate overall performance.
\begin{figure}
    \centering
    \includegraphics[width=0.9\linewidth]{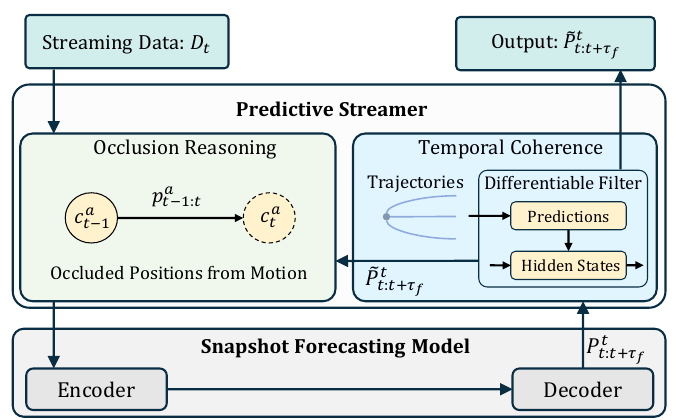}
    \vspace{-2mm}
    \caption{Predictive streamer. It serves as an intermediate integration between snapshot-based forecasters and streaming data. The occlusion reasoning module estimates the positions of occluded agents from the predicted movement (Sec.~\ref{sec:occlusion_reasoning}). Differentiable filters (DFs) estimate future trajectories as their hidden states and refine the forecasting results through temporal coherence in adjacent frames (Sec.~\ref{sec:df}).}
    \vspace{-2mm}
    \label{fig:system}
\end{figure}
\section{ALGORITHMS}
\label{sec:algo}

\subsection{Pipeline of Streaming Forecasting}
\label{sec:adaptation}

We propose a general pipeline to adapt a well-learned snapshot-based model seamlessly into streaming forecasting.

\mypar{Three-stage training.} (1) \textit{Pretraining}. We train a standard snapshot-based model on the forecasting dataset of Argoverse to reasonably initialize the forecaster. (2) \emph{Finetuning}. We extract snapshots from the Argoverse-SF training set to finetune the pretrained model. This mitigates the distribution shift between Argoverse's forecasting dataset and Argoverse-SF. (3) \emph{Streaming training}. We design and train \emph{predictive streamer} to better integrate with the streaming setup, detailed in Sec.~\ref{sec:occlusion_reasoning} and Sec.~\ref{sec:df}.

\mypar{Streaming inference.} To align with the interface of a snapshot-based forecaster, we integrate it into streaming data in a sliding window way. At frame $t$, the model takes as input the last $\tau_h$ frames, $D_{t-\tau_h+1:t}$, and predicts future $\tau_f$ frames, $P_{t:t+\tau_f}$. If our predictive streamer is available, it uses predictions and agent features from the forecaster and refines $P_{t:t+\tau_f}$ into $\widetilde{P}_{t:t+\tau_f}$ as final predictions for output.

\mypar{Baseline.} Hallucinating the occluded positions is necessary to operate forecasting models on all the agents. Our baseline uses a Kalman filter to predict occluded positions, inspired by how 3D multi-object tracking addresses occlusions~\cite{simpletrack}\cite{wang2021immortal}. 

\mypar{Overview of predictive streamers.} Occlusion reasoning and temporal coherence are emergent challenges in streaming forecasting. The baseline described above lacks specialized solutions to address these issues. Therefore, our \emph{predictive streamer} accommodates these challenges and acts as a general intermediate component as in Fig.~\ref{fig:system}, better integrating a snapshot-based forecaster into the streaming setup.

\begin{algorithm}[t]
\caption{Algorithm for Occlusion Reasoning}
\begin{algorithmic}[1]
\label{algo:occlusion_reasoning}
\renewcommand{\algorithmicrequire}{\textbf{Input:}}
\renewcommand{\algorithmicensure}{\textbf{Output:}}
\REQUIRE \ \\
$D_{1:t}$: The input data, $D_t=\{(c_t^a, v_t^a)\}_{a\in A_t}$ are coordinates and visibility masks (see Sec.~\ref{sec:streaming_formulation}). \\
$P_{t-1:t+\tau_f-1}$: Predictions of $\mathtt{Decoder}$ at frame $t-1$. \\
$M$: HD-Maps.
\ENSURE \ \\ $\widetilde{P}_{t:t+\tau_f}$: Predictions at frame $t$.
\\ \vspace{-2mm} \hrulefill
\FOR{arbitrary agent $a\in A_t$}
\IF{$v^a_t$ is false ($a$ is invisible)}
\STATE Choose the most confident $\widehat{p}^a_{t:t+\tau_f}$ from $p^a_{t:t+\tau_f}$
\STATE $c_t^a \xleftarrow{} c_{t-1}^a + \widehat{p}^a_{t-1:t}$
\ENDIF
\ENDFOR
\STATE Encode agent features: $F_t\xleftarrow{} \mathtt{Encoder}(D_{t-\tau_h+1:t}, M)$
\STATE Multi-future forecasting: $P_{t:t+\tau_f}\xleftarrow{} \mathtt{Decoder}(F_t)$
\STATE $P_{t:t+\tau_f}$ acts as the final result: $\widetilde{P}_{t:t+\tau_f}\xleftarrow{}P_{t:t+\tau_f}$
\RETURN $\widetilde{P}_{t:t+\tau_f}$
\end{algorithmic}
\end{algorithm}

\subsection{Forecasting for Occluded Agents}
\label{sec:occlusion_reasoning}

\mypar{Occlusion reasoning with multi-modal predictions.} Estimating the occluded positions is the prerequisite of forecasting for those agents. Its common approach is adding predicted motion to previously visible positions. Although our Kalman filter baseline (Sec.~\ref{sec:adaptation}) follows this intuition, it is insufficient in that the Kalman filter is unable to handle complex motions and contextual information, such as HD-Maps and the movements of surrounding agents. Therefore, we propose to leverage the forecasting results from snapshot-based models to provide more accurate estimation. In practice, a special branch $\widehat{p}_{t:t+\tau_f}^t$ from multi-modal trajectories is selected according to the highest confidence scores. The movement in $\widehat{p}_{t:t+\tau_f}^t$ propagates the agents' positions through the occluded timestamp as in Algorithm~\ref{algo:occlusion_reasoning}. Specifically, if an agent $a$ is occluded, we replace its invisible location with previous motion $c_{t-1}^a + \widehat{p}^a_{t-1:t}$. We highlight that the selection process connects the multi-modal trajectories with the single-modal input of forecasters. In this way, our approach retains the advantage of multi-modal trajectories from learning-based forecasters, including their diverse movements, HD-Map context, and cross-agent interaction.

\mypar{Discussion.} Our occlusion reasoning is different from prior 3D perception work \cite{pang2023standing}\cite{simpletrack}\cite{wang2021immortal} under a forecasting context. First, we leverage \emph{multi-modal} trajectories, which cover a larger spectrum of distributions and thus outperform single-modal trajectories (empirically validated in Sec.~\ref{sec:quantitative}). Second, while previous studies avoid updating predictions on occluded frames due to unreliable information, our streaming module updates $p_{t:t+\tau_f}$ on every timestamp, providing a solution for handling long occlusions ($>\tau_f$ frames) and incorporating the latest information for instant reactions. The remaining unreliability challenge is addressed by leveraging temporal coherence, which is discussed in Sec.~\ref{sec:df}.

\subsection{Differentiable Filters for Temporal Coherence}
\label{sec:df}

\begin{figure}
    \centering
    \includegraphics[width=0.9\linewidth]{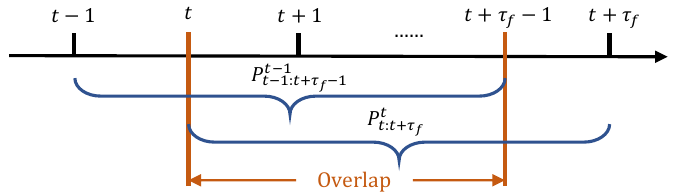}
    \vspace{-2mm}
    \caption{Illustration of temporal coherence. The predictions on adjacent frames ($t-1$ and $t$) have a large overlap, and such consistency is useful for improving the forecasting quality.}
    \vspace{-4mm}
    \label{fig:coherence}
\end{figure}

\mypar{Temporal coherence.} As streaming forecasting operates on continuous frames, adjacent timestamps have strong temporal consistency because of their large overlaps, as in Fig.~\ref{fig:coherence}. Suppose we use the superscript in $\widetilde{p}_{t:t+\tau_f}^t$ to denote predictions originating from $t$. The trajectories on $t-1$, denoted as $\widetilde{p}_{t-1:t+\tau_f-1}^{t-1}$, and the trajectories on $t$, denoted as $\widetilde{p}_{t:t+\tau_f}^{t}$, should be aligned between frames $t$ and $t+\tau_f-1$, which cover most of their prediction horizons. To account for the predictions in nearby frames, \emph{our key insight} is that differentiable filters (DFs) can effectively enhance temporal coherence by recursively fusing predictions into their hidden states. 

\mypar{Background of DFs.} The basis of DFs is Bayes filtering, \emph{e.g.}, Kalman filters~\cite{kalman1960new}. It considers a \emph{process model} and an \emph{observation model} as Eqn.~\ref{eqn:dynamics}, where $x_t$ denotes hidden states, $z_t$ denotes observations and the assumptions are linear dynamics, no control signals, and Gaussian distribution. Additionally, $w$ and $\delta$ are noises following distributions $N(0, Q)$ and $N(0, R)$, where $Q$ and $R$ are covariance matrices.
\begin{equation}
    \label{eqn:dynamics}
    x_{t}=A x_{t-1} + w,\ \text{and}\ z_t=C x_t + \delta.
\end{equation}
The filter recursively approximates the posterior distribution of $x_t$ in the form of $N(\mu_{x_t}, \Sigma_{x_t})$. During this process, DF~\cite{haarnoja2016backprop} proposes using a neural network $\phi$ to infer low-dimensional $z_t$ from high-dimensional sensor inputs or generate adaptive covariance $R$. As the Kalman filter is a differentiable process, the neural network $\phi$ can be learned jointly with it.

\mypar{DFs for motion forecasting.} In the case of motion forecasting, we treat future trajectories $p_{t:t+\tau_f}^t$ as both the hidden states $x_t$ and observation $z_t$ in DF. Then DF represents the hidden states with a Gaussian distribution $N(\mu_{x_t}, \Sigma_{x_t})$ and recursively estimates the mean $\mu_{x_t}$ of the hidden states $x_t$ as refined forecasting results, denoted as $\widetilde{p}_{t:t+\tau_f}^{t}$. We illustrate this process as follows:
\begin{equation}
    \label{eqn:df_interface}
    \mu_{x_t} \xleftarrow{} \mathtt{DF}(p^{t}_{t:t+\tau_f}, \mu_{x_{t-1}}, \Sigma_{x_{t-1}}), \ \text{then}\ \widetilde{p}^{t}_{t:t+\tau_f} \xleftarrow{} \mu_{x_t}.
\end{equation}

\begin{figure}
    \centering
    \includegraphics[width=0.9\linewidth]{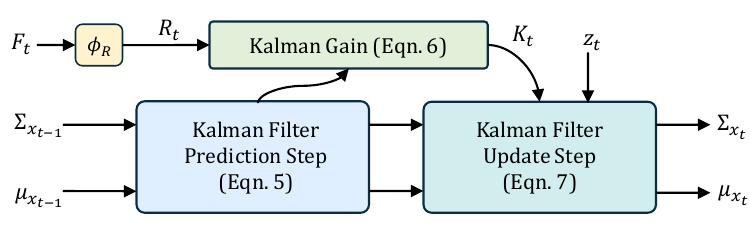}
    \vspace{-2mm}
    \caption{Computation of differentiable filters. A neural network $\phi_R$ predicts the observation covariance $R_t$ from agent features. Then a Kalman filter fuses the latest predictions into the hidden states accordingly.}
    \vspace{-2mm}
    \label{fig:df}
\end{figure}

In DF, we first use a learnable neural network $\phi_R$ to predict adaptive covariance $R_t$ for varied agents and timestamps based on their features $F_t$, which enables DF to fuse the highly stochastic trajectories properly. Then we feed the predicted $R_t$ into a standard Kalman filter to update $\mu_{x_{t-1}}$ as in Fig.~\ref{fig:df}, according to the following filtering steps.
\begin{align}
    \label{eqn:kf_predict}
    \widetilde{\mu}_{x_t} \xleftarrow{} A \mu_{x_{t-1}},\ \widetilde{\Sigma}_{x_t} \xleftarrow{} A \Sigma_{x_{t-1}} A^{\top} + Q_t\  \text{[Prediction]}& \\
    \label{eqn:kalman_gain}
    K_t \xleftarrow{} \widetilde{\Sigma}_{x_t} C^\top (C \widetilde{\Sigma}_{x_t} C^\top + R_t)^{-1}\ \text{[Kalman gain]}& \\
    \label{eqn:kf_update}
    \mu_{x_t}\xleftarrow{}\widetilde{\mu}_{x_t}+K_t(z_t-C \widetilde{\mu}_{x_t}), \Sigma_{x_t}\xleftarrow{} (\mathbb{I}-K_tC)\widetilde{\Sigma}_{x_t}\ \text{[Update]}&
\end{align}

We further discover that \emph{specifying inter-frame dynamics} is the key to capturing temporal coherence. Taking the movements on X-axis for example, denoted as $\widetilde{p}^{t,x}_{t:t+\tau_f}$, we define the matrix $A$ in the process model of DF (Eqn.~\ref{eqn:dynamics}) as:
\begin{equation}
\label{eqn:df_dynamics}
A \xleftarrow{} \left[ \begin{array}{c|c}
 0_{(\tau_f-1)\times 1} & \mathbb{I}_{(\tau_f-1)\times(\tau_f-1)} \\ \hline
  0_{1\times 1} & 1_{1\times 1}
\end{array}\right],\
    \widetilde{p}^{t,x}_{t:t+\tau_f} \xleftarrow{}
A \widetilde{p}^{t-1,x}_{t-1:t+\tau_f-1}.
\end{equation}
The above matrix $A$ uses the identity matrix $\mathbb{I}_{(\tau_f-1)\times(\tau_f-1)}$ to indicate that the overlapped forecasting frames, \emph{i.e.}, $\widetilde{p}^{t,x}_{t:t+\tau_f-1}$ and $\widetilde{p}^{t-1,x}_{t:t+\tau_f-1}$, should be consistent. The bottom-right $1_{1\times 1}$ means that the process model pads the additional timestamp $t+\tau_f$ with the last movement on the previous timestamp, where we assume that motion cannot change drastically. As for $C$ in the observation model, we define it as an identity matrix $\mathbb{I}$, because the hidden states $x_t$ and observations $z_t$ both represent future movements.

For implementation, we follow prior work~\cite{haarnoja2016backprop}\cite{kloss2021train} in treating $Q_t$ as a fixed hyper-parameter. Moreover, as our future trajectories have larger dimensions than conventional state estimation tasks, we assume $R_t$ is diagonal and predict it by $(\phi_R(F_t))^2$ to avoid invalid covariance matrices.

\mypar{Training and inference.} We train DF using the same forecasting loss penalizing the distance between predictions and ground truth (details in Sec.~\ref{sec:implementations}). During inference, we recursively apply DF as in Fig.~\ref{fig:df} to refine every trajectory. 

\mypar{Discussion.} Our DF in predictive streamer not only improves trajectory prediction but also highlights the temporal coherence property of streaming forecasting. In response to the question from the discussion of Sec.~\ref{sec:occlusion_reasoning}, our DF-based approach allows the occluded frames to aggregate the information from previous visible frames, which addresses the unreliability issue of hallucinated occluded positions.
\section{EXPERIMENTS}

\subsection{Implementation Details}
\label{sec:implementations}

\mypar{Snapshot forecasting models.} We use two encoder/decoder architectures for generalizability: VectorNet~\cite{gao2020vectornet} and mmTransformer~\cite{liu2021multimodal}. 
To ensure a fair comparison, we \emph{double the number of layers} of the original VectorNet, as it was proposed earlier and relatively light-weight. Our loss functions follow the \emph{winner-takes-all} strategy, where the best branch is selected from the multi-modal trajectories based on the smallest distance to ground truth. Then a cross-entropy loss supervises its confidence score to approach 1, and a smooth L1 loss penalizes its distance to the ground truth. Our re-implementation of VectorNet and mmTransformer outperforms their reported results in~\cite{gao2020vectornet}\cite{liu2021multimodal}.

\mypar{Differentiable filters.} $\phi_R$ adopts the same multi-layer perceptron (MLP) architecture as the decoder in VectorNet. During streaming training, we freeze the encoder and decoder of the snapshot-based forecaster and supervise DFs with the same smooth L1 loss for multi-modal trajectories.

\mypar{Model training.} We introduce the details of three-stage training. (1) \emph{Pretraining}: we use AdamW~\cite{loshchilov2017decoupled}, 5e-4 as learning rate, 1e-4 as weight decay, and train for 24 epochs. (2) \emph{Finetuning}: we use AdamW, 1e-4 as learning rate, 1e-4 as weight decay, and train for 8 epochs. The training samples come from Argoverse-SF, and every iteration involves \emph{forecasting on 5 adjacent frames}. (3) \emph{Streaming training}: we freeze the snapshot encoder and decoder and adopt the same configuration as the finetuning stage.

\begin{table*}
\vspace{1mm}
\caption{Analysis of predictive streamer with VectorNet. We analyze the overall metrics (primary) and the breakdown metrics specifically for moving agents (secondary). Beginning from the baseline of directly adapting snapshot models with a Kalman filter (Sec.~\ref{sec:adaptation}), we use the most confident trajectory for occlusion reasoning (``OCC'') and learn a differentiable filter to improve the temporal coherence (``DF''). Our strategies significantly improve the performance, indicating the benefits of viewing motion forecasting from a streaming perspective. More analysis is in Sec.~\ref{sec:ablation}.}\vspace{-2mm}
\begin{subtable}[h]{0.49\linewidth}
\caption{Evaluation metrics (K=6) with VectorNet.}\vspace{-2mm}
\centering
\resizebox{1.0\linewidth}{!}
{
        \begin{tabular}{
        c@{\hspace{0.8mm}}c@{\hspace{0.8mm}}|
        c@{\hspace{1.0mm}}c@{\hspace{1.0mm}}c@{\hspace{1.0mm}}|
        c@{\hspace{1.0mm}}c@{\hspace{1.0mm}}c@{\hspace{1.0mm}}|
        c@{\hspace{1.0mm}}c@{\hspace{1.0mm}}c@{\hspace{1.0mm}}}
        \toprule
        \multirow{2}{*}{OCC} & \multirow{2}{*}{DF} & \multicolumn{3}{c|}{Overall} & \multicolumn{3}{c|}{Moving-Occluded} & \multicolumn{3}{c}{Moving-Visible}  \\
         & & minFDE$\downarrow$ & minADE$\downarrow$ & MR$\downarrow$ & minFDE$\downarrow$ & minADE$\downarrow$ & MR$\downarrow$ & minFDE$\downarrow$ & minADE$\downarrow$ & MR$\downarrow$  \\
        \midrule
        & & 1.67 & 1.64 & 0.20 & 4.05 & 4.59 & 0.50 & 1.70 & 1.09 & 0.25  \\
        \cmark & & 1.43	& 1.24 & 0.18 & 3.22 & 3.17 & 0.44 & 1.70 & 1.09 & 0.25  \\
        \cmark & \cmark & \textbf{1.37} & \textbf{1.20} & \textbf{0.17} & \textbf{3.03} & \textbf{3.07} & \textbf{0.41} & \textbf{1.67} & \textbf{1.07} & \textbf{0.24}  \\
        \bottomrule
       \end{tabular}
}
\label{tab:vectornet_ablation_k6}
\end{subtable}
\begin{subtable}[h]{0.49\linewidth}
\caption{Evaluation metrics (K=1) with VectorNet.}\vspace{-2mm}
\centering
\resizebox{1.0\linewidth}{!}
{
        \begin{tabular}{
        c@{\hspace{1.0mm}}c@{\hspace{1.0mm}}|
        c@{\hspace{1.0mm}}c@{\hspace{1.0mm}}c@{\hspace{1.0mm}}|
        c@{\hspace{1.0mm}}c@{\hspace{1.0mm}}c@{\hspace{1.0mm}}|
        c@{\hspace{1.0mm}}c@{\hspace{1.0mm}}c@{\hspace{1.0mm}}}
        \toprule
        \multirow{2}{*}{OCC} & \multirow{2}{*}{DF} & \multicolumn{3}{c|}{Overall} & \multicolumn{3}{c|}{Moving-Occluded} & \multicolumn{3}{c}{Moving-Visible}  \\
         & & minFDE$\downarrow$ & minADE$\downarrow$ & MR$\downarrow$ & minFDE$\downarrow$ & minADE$\downarrow$ & MR$\downarrow$ & minFDE$\downarrow$ & minADE$\downarrow$ & MR$\downarrow$  \\
        \midrule
        & & 3.90 & 2.57 & 0.37 & 9.39 & 7.23 & \textbf{0.75}	& 4.63 & 1.94 & 0.63  \\
        \cmark & & 3.42	& 2.05 & \textbf{0.36} & 7.84 & 5.42 & \textbf{0.75} & 4.63 & 1.94 & 0.63 \\
        \cmark & \cmark & \textbf{3.32} & \textbf{1.98} & \textbf{0.36} & \textbf{7.42} & \textbf{5.16} & 0.76 & \textbf{4.57} & \textbf{1.89} & \textbf{0.62}  \\
        \bottomrule
       \end{tabular}
}
\label{tab:vectornet_ablation_k1}
\end{subtable}
\label{tab:vectornet_ablation}
\end{table*}
\begin{table*}
\caption{Analysis of predictive streamer with mmTransformer. Abbreviations are the same as Table~\ref{tab:vectornet_ablation}. Our algorithm improves the forecasting quality on mmTransformer, further supporting the significance of exploring streaming forecasting. }\vspace{-2mm}
\begin{subtable}[h]{0.49\linewidth}
\caption{Evaluation metrics (K=6) with mmTransformer.}\vspace{-2mm}
\centering
\resizebox{1.0\linewidth}{!}
{
        \begin{tabular}{
        c@{\hspace{1.0mm}}c@{\hspace{1.0mm}}|
        c@{\hspace{1.0mm}}c@{\hspace{1.0mm}}c@{\hspace{1.0mm}}|
        c@{\hspace{1.0mm}}c@{\hspace{1.0mm}}c@{\hspace{1.0mm}}|
        c@{\hspace{1.0mm}}c@{\hspace{1.0mm}}c@{\hspace{1.0mm}}}
        \toprule
        \multirow{2}{*}{OCC} & \multirow{2}{*}{DF} & \multicolumn{3}{c|}{Overall} & \multicolumn{3}{c|}{Moving-Occluded} & \multicolumn{3}{c}{Moving-Visible}  \\
         & & minFDE$\downarrow$ & minADE$\downarrow$ & MR$\downarrow$ & minFDE$\downarrow$ & minADE$\downarrow$ & MR$\downarrow$ & minFDE$\downarrow$ & minADE$\downarrow$ & MR$\downarrow$  \\
        \midrule
        & & 1.74 & 1.61 & 0.19 & 4.27 & 4.48 & 0.49 & 1.70 & 1.07 & 0.23  \\
        \cmark & &  1.61 & 1.37 & 0.18 & 3.89 & 3.67 & 0.43 & 1.70 & 1.07 & 0.23 \\
        \cmark & \cmark & \textbf{1.54} & \textbf{1.30} & \textbf{0.17} & \textbf{3.60} & \textbf{3.40} & \textbf{0.41} & \textbf{1.66} & \textbf{1.05} & 0.23 \\
        \bottomrule
       \end{tabular}
}
\label{tab:mmtrans_ablation_k6}
\end{subtable}
\begin{subtable}[h]{0.49\linewidth}
\caption{Evaluation metrics (K=1) with mmTransformer.}\vspace{-2mm}
\centering
\resizebox{1.0\linewidth}{!}
{
        \begin{tabular}{
        c@{\hspace{1.0mm}}c@{\hspace{1.0mm}}|
        c@{\hspace{1.0mm}}c@{\hspace{1.0mm}}c@{\hspace{1.0mm}}|
        c@{\hspace{1.0mm}}c@{\hspace{1.0mm}}c@{\hspace{1.0mm}}|
        c@{\hspace{1.0mm}}c@{\hspace{1.0mm}}c@{\hspace{1.0mm}}}
        \toprule
        \multirow{2}{*}{OCC} & \multirow{2}{*}{DF} & \multicolumn{3}{c|}{Overall} & \multicolumn{3}{c|}{Moving-Occluded} & \multicolumn{3}{c}{Moving-Visible}  \\
         & & minFDE$\downarrow$ & minADE$\downarrow$ & MR$\downarrow$ & minFDE$\downarrow$ & minADE$\downarrow$ & MR$\downarrow$ & minFDE$\downarrow$ & minADE$\downarrow$ & MR$\downarrow$  \\
        \midrule
        & & 3.78 & 2.44 & 0.38 & 9.25 & 6.83 & 0.80 & 4.25 & 1.81 & 0.59  \\
        \cmark & & 3.30 & 1.98 & \textbf{0.35} & 7.42 & 5.20 & \textbf{0.77} & 4.24 & 1.80 & 0.58 \\
        \cmark & \cmark & \textbf{3.25} & \textbf{1.93} & \textbf{0.35} & \textbf{7.21} & \textbf{4.94} & \textbf{0.77} & \textbf{4.22} & \textbf{1.78} & \textbf{0.57} \\
        \bottomrule
       \end{tabular}
}
\label{tab:mmtrans_ablation_k1}
\end{subtable}
\label{tab:mmtrans_ablation}
\end{table*}

\subsection{Ablation Studies}
\label{sec:ablation}

\mypar{Predictive streamer.} We analyze \emph{predictive streamer} in Table~\ref{tab:vectornet_ablation} and Table~\ref{tab:mmtrans_ablation} based on the encoders and decoders from VectorNet and mmTransformer, respectively. As illustrated, \emph{every streaming strategy significantly improves the forecasting quality}, especially for the minFDE and minADE. We highlight that DF is vital for better predictions. It refines occlusion reasoning by fusing the results from previous reliable frames. Furthermore, better temporal coherence also improves the trajectories of visible agents without modifying the underlying encoders and decoders, which is plug-and-play for deploying snapshot forecasters.

\mypar{Properties of streaming datasets.} Before further analysis, we clarify the differences between a snapshot-based dataset and our streaming datasets, which are reflected in Table~\ref{tab:vectornet_ablation} and Table~\ref{tab:mmtrans_ablation} and subtle to notice at first look. (1) The minADE for occluded cases can be larger than minFDE. This is because minADE computes more occlusion cases than minFDE in Argoverse-SF. Concretely, we adopt a ``masked'' minADE that computes the metrics once there is one frame with ground truth (as in Sec.~\ref{sec:benchmark}, evaluation metrics); thus, any prediction accounted by minFDE is also considered by minADE, but not the reverse. (2) The static agents (excluded in the table due to space limit) have much smaller minADE and minFDE values and decrease the scales of changes in overall metrics. (3) We compare the performance between VectorNet and mmTransformer, where mmTransformer is better for visible agents and K=1 cases, but fails in the K=6 occlusion scenarios. This is because the queries of mmTransformer are tailored to distinct movement patterns, and their less confident directions have a weaker ability to adapt to noisy occlusions. Even so, our predictive streamer consistently improves for all scenarios.

\mypar{Three-stage training.} The three-stage procedure of adapting forecasters, which is proposed in Sec.~\ref{sec:adaptation}, is necessary: finetuning improves the performance, where the overall minFDE decreases from 3.07m to 1.67m and the overall MR drops from 25\% to 20\%.

\subsection{Quantitative Analysis}
\label{sec:quantitative}

We conduct several analyses for our design choices and provide further insights for streaming forecasting. By default, we use the VectorNet encoder and decoder.

\begin{table}[tb]
\caption{Comparing DF with Kalman filter (KF) and LSTM for temporal coherence. Our DF outperforms both KF and LSTM, especially for occlusion cases.}\vspace{-2mm}
\begin{subtable}[h]{1.0\linewidth}
\caption{Overall metrics.}\vspace{-1mm}
\centering
    \resizebox{0.95\linewidth}{!}{
    \begin{tabular}{c|ccc|ccc}
    \toprule
        \multirow{2}{*}{Model} & \multicolumn{3}{c|}{Overall (K=6)} & \multicolumn{3}{c}{Overall (K=1)} \\
        & minFDE$\downarrow$ & minADE$\downarrow$ & MR$\downarrow$ & minFDE$\downarrow$ & minADE$\downarrow$ & MR$\downarrow$  \\
        \midrule
        Snapshot & 1.67 & 1.64 & 0.20 & 3.90 & 2.57 & 0.37  \\
        KF & 1.42 & 1.24 & 0.18 & 3.39 & 2.03 & 0.37 \\
        LSTM & 1.56 & 1.39 & 0.18 & 3.75 & 2.30 & \textbf{0.36} \\
        DF (Ours) & \textbf{1.37} & \textbf{1.20} & \textbf{0.17} & \textbf{3.32} & \textbf{1.98} & \textbf{0.36}   \\
        \bottomrule
    \end{tabular}
    }
    \label{tab:recurrent_overall}
\end{subtable}
\begin{subtable}[h]{1.0\linewidth}
\caption{Breakdown metrics.}\vspace{-2mm}
\centering
    \resizebox{0.95\linewidth}{!}{
    \begin{tabular}{c|ccc|ccc}
    \toprule
        \multirow{2}{*}{Model} & \multicolumn{3}{c|}{Moving-Occluded (K=6)} & \multicolumn{3}{c}{Moving-Visible (K=6)} \\
        & minFDE$\downarrow$ & minADE$\downarrow$ & MR$\downarrow$ & minFDE$\downarrow$ & minADE$\downarrow$ & MR$\downarrow$  \\
        \midrule
        Snapshot & 4.05 & 4.59 & 0.50 & 1.70 & 1.09 & 0.25  \\
        KF & 3.13 & 3.15 & 0.42 & 1.70 & 1.09 & 0.25 \\
        LSTM & 3.74 & 3.78 & 0.46 & 1.68 & \textbf{1.07} & 0.25  \\
        DF (Ours) & \textbf{3.03} & \textbf{3.07} & \textbf{0.41} & \textbf{1.67} & \textbf{1.07} & \textbf{0.24}   \\
        \bottomrule
    \end{tabular}
    }
    \label{tab:recurrent_breakdown}
\end{subtable}
\vspace{-2mm}
    \label{tab:recurrent}
\end{table}
\mypar{Comparing DF with KF and LSTM.} We compare the effects of DF and LSTM, because LSTM is a common approach for temporal coherence~\cite{haarnoja2016backprop}\cite{kloss2021train}. Specifically, we replace the DF in our predictive streamer with a 4-layer LSTM to enhance $F_t$, and jointly train it with the decoder. For a fair comparison, both DF and LSTM \emph{adopt the occlusion reasoning} described in Sec.~\ref{sec:occlusion_reasoning}. As shown in Table~\ref{tab:recurrent_overall}, LSTM cannot outperform our DF. After more detailed analysis in Table~\ref{tab:recurrent_breakdown}, we find that LSTM can slightly improve over the snapshot models on visible cases, but not as large as DFs, because DF explicitly specifies temporal coherence by a process model (Eqn.~\ref{eqn:df_dynamics}). The main disadvantage of LSTMs is on the occluded cases because the problematic features for occluded agents can corrupt the hidden states of LSTM and influence the features in subsequent frames. We also compare DF to the baseline of Kalman filters, where DF has a better performance by learning adaptive covariance matrices.

\begin{figure}[tb]
    \centering
    \includegraphics[width=0.90\linewidth]{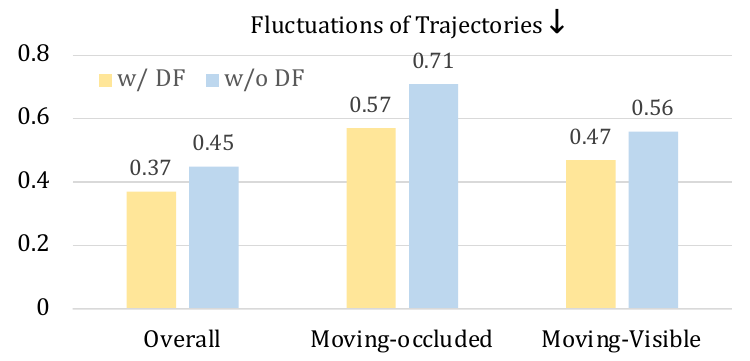}
    \vspace{-2mm}
    \caption{DF decreases the fluctuations of trajectories by 10\%-20\%. Y-axis is measured by meters/frame.}
    \vspace{-4mm}
    \label{fig:smoothness}
\end{figure}
\mypar{DF improves temporal coherence.} 
We mimic~\cite{kanazawa2019learning} in defining a second-order metric ``fluctuation'' to analyze the temporal coherence. Specifically, we convert predictions on $t-1$ and $t$ into their absolute coordinates $\widehat{c}_{t-1:t+\tau_f-1}^{t-1}$ and $\widehat{c}_{t:t+\tau_f}^{t}$ and compute their distance in overlapped frames $\frac{1}{\tau_f-1}\|\widetilde{c}_{t:t+\tau_f-1}^{t-1} - \widetilde{c}_{t:t+\tau_f-1}^{t} \|_2$. As in Fig.~\ref{fig:smoothness}, DF decreases the fluctuation by a large margin.

\mypar{Occlusion reasoning is better with multi-modality.} We compare our estimation of occluded positions with previous 3D perception strategies that capitalize occlusion reasoning~\cite{simpletrack}\cite{pang2023standing}\cite{wang2021immortal}. The main distinction is that we utilize multi-modal predictions. To validate the benefits of multi-modality, we integrate an additional single-modal decoder $\mathtt{Decoder}_O$ and a DF into our predictive streamer for occlusion reasoning. As in Table~\ref{tab:occlusion_reasoning}, \emph{multi-modal predictions are beneficial to estimating invisible positions}. The reason is that multi-modal trajectories can cover a larger spectrum of futures. We also notice that the improvement of DF generalizes to both single-modal and multi-modal trajectories. 
\begin{table}[tb]
\caption{Multi-modality trajectories are better for occlusion reasoning. The metrics are evaluated under K=6. Using DF also enhances the performance for both single-modal and multi-modal predictions.}\vspace{-2mm}
    \centering
    \resizebox{1.0\linewidth}{!}{
    \begin{tabular}{c|c|c@{\hspace{1.0mm}}c@{\hspace{1.0mm}}c@{\hspace{1.0mm}}|c@{\hspace{1.0mm}}c@{\hspace{1.0mm}}c@{\hspace{1.0mm}}}
    \toprule
        \multirow{2}{*}{Modality} & \multirow{2}{*}{DF} & \multicolumn{3}{c|}{Overall} & \multicolumn{3}{c}{Moving-Occluded} \\
        & & minFDE$\downarrow$ & minADE$\downarrow$ & MR$\downarrow$ & minFDE$\downarrow$ & minADE$\downarrow$ & MR$\downarrow$  \\
        \midrule
        \multirow{2}{*}{Single} & \xmark & 1.53 & 1.36 & \textbf{0.17} & 3.54 & 3.59 & 0.42 \\
        & \cmark & 1.40 & 1.25 & \textbf{0.17} & 3.14 & 3.20 & \textbf{0.41} \\
        \midrule
         \multirow{2}{*}{Multiple} & \xmark & 1.43 & 1.24 & 0.18 & 3.22 & 3.17 & 0.44 \\
        & \cmark & \textbf{1.37} & \textbf{1.20} & \textbf{0.17} & \textbf{3.03} & \textbf{3.07} & \textbf{0.41} \\
        \bottomrule
    \end{tabular}
    }
    \vspace{-3mm}
    \label{tab:occlusion_reasoning}
\end{table}

\subsection{Qualitative Analysis}

\begin{figure}[tb]
    \centering
    \includegraphics[width=1.0\linewidth]{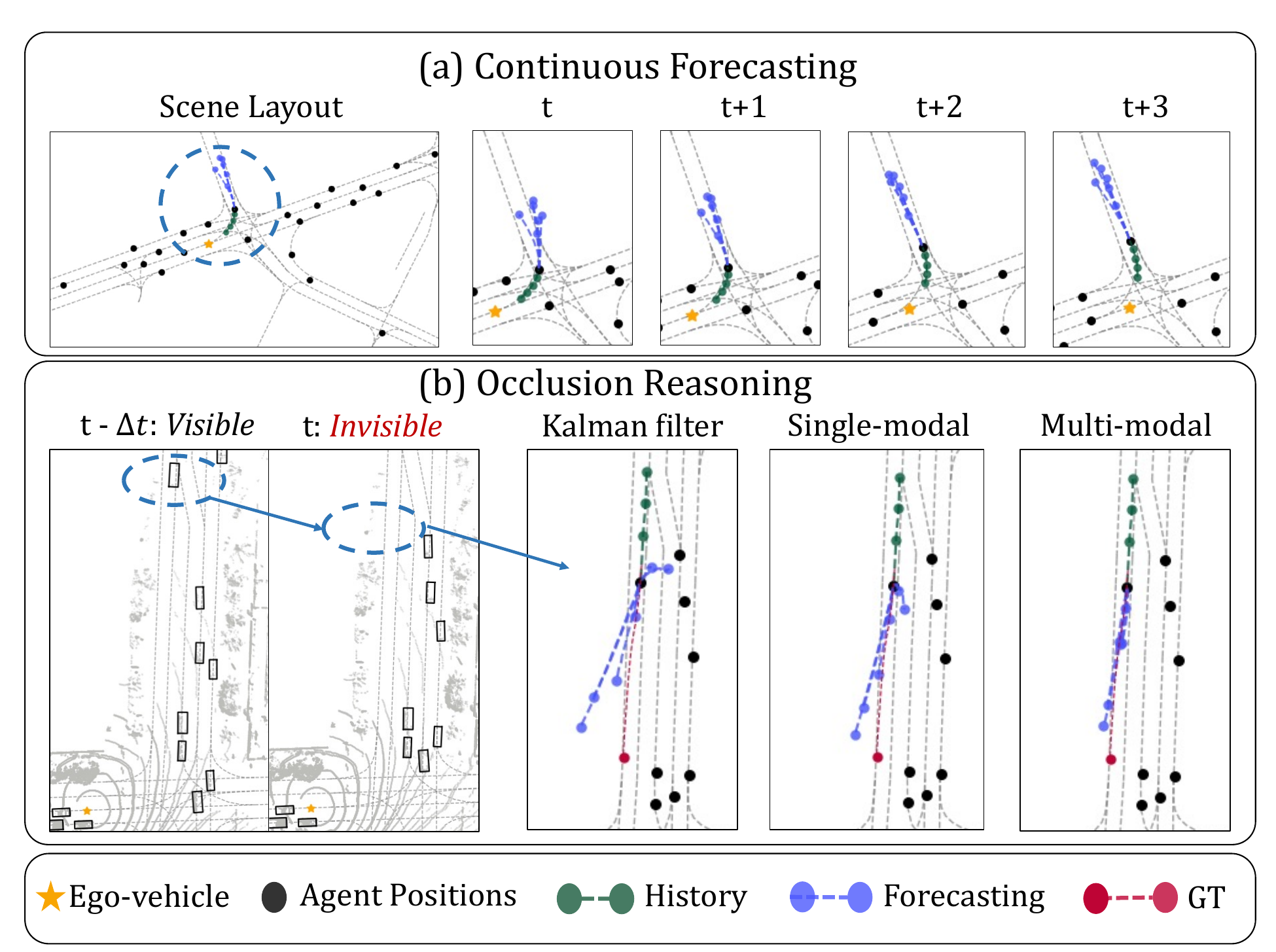}
    \caption{(\textbf{Best viewed in color}) We visualize the streaming forecasting results. Only the interesting agent is visualized for clarity. (a) When forecasting proceeds in continuous frames, the trajectories (blue dashed lines) gradually adapt and change according to the latest agent positions. (b) To deal with an occluded agent, using multi-modal trajectories is better than the conventional Kalman filters or single-modal trajectories, because the predicted trajectories (blue dashed lines) are closer to the ground truth (red lines). Please note that point clouds are only for visualization purposes.}
    \vspace{-4mm}
    \label{fig:visualization}
\end{figure}

We first illustrate the intuition of streaming forecasting in Fig.~\ref{fig:visualization}\textcolor{red}{a}, where our predictions gradually adapt to an agent's incoming observations. For example, the trajectories have smaller divergence on $t+3$ than on $t$. Furthermore, we demonstrate the challenge of occlusion reasoning in Fig.~\ref{fig:visualization}\textcolor{red}{b}. Although the errors for occluded positions are subtle, the Kalman filter and single-modal methods have seriously wrong predictions, while our multi-modal method has relatively larger fidelity. Please check our demo for more visualizations.

\section{CONCLUSIONS AND FUTURE WORK}

This paper proposes a novel perspective to study motion forecasting: \emph{streaming forecasting}. By repurposing tracking data, we capture the intrinsic challenges of \emph{forecasting for occluded agents} and \emph{temporal coherence} in real-world traffic, which are overlooked by previous snapshot-based setups. In addition, we propose general \emph{predictive streamers} that leverage multi-modal forecasting to address occlusions and introduce differentiable filters to enhance temporal continuity. These solutions and analyses intend to generate interest in studying motion forecasting in a realistic streaming setting.

As the first study on streaming forecasting, we expect future works to improve predictive streamers for miss rates. In addition, the benchmark is also extensible to more types of agents and larger datasets.

\noindent{\textbf{Acknowledgement.} This work was supported in part by NSF Grant 2106825, NIFA Award 2020-67021-32799, the Jump ARCHES endowment, the NCSA Fellows program, the IBM-Illinois Discovery Accelerator Institute, the Illinois-Insper Partnership, and the Amazon Research Award.} 

\bibliographystyle{IEEEtranS}
\bibliography{IEEEabrv,reference}

\end{document}